\documentclass[10pt,twocolumn,letterpaper]{article}

\usepackage[pagenumbers]{cvpr} 
\usepackage{pifont}
\usepackage{graphicx}
\usepackage{overpic}
\usepackage{multirow}
\usepackage{marvosym}

\definecolor{cvprblue}{rgb}{0.21,0.49,0.74}
\usepackage[pagebackref,breaklinks,colorlinks,allcolors=cvprblue]{hyperref}

\def\paperID{1080} 
\def\confName{CVPR}
\def\confYear{2025}

\title{OmniGuard: Hybrid Manipulation Localization via Augmented \\ Versatile Deep Image Watermarking}

\author{
\href{https://xuanyuzhang21.github.io/}{Xuanyu Zhang}\textsuperscript{1,2 *}, Zecheng Tang\textsuperscript{1 *}, Zhipei Xu\textsuperscript{1 *}, Runyi Li\textsuperscript{1}, Youmin Xu\textsuperscript{1}, \\ Bin Chen\textsuperscript{1}, Feng Gao\textsuperscript{3 \Letter}, \href{https://jianzhang.tech/}{Jian Zhang}\textsuperscript{1,2 \Letter}\\
\textsuperscript{1} School of Electronic and Computer Engineering, Peking University\\
\textsuperscript{2} Peking University Shenzhen Graduate School-Rabbitpre AIGC Joint Research Laboratory \\
\textsuperscript{3} School of Arts, Peking University
}

\begin{document}


\maketitle
\begin{abstract}
With the rapid growth of generative AI and its widespread application in image editing, new risks have emerged regarding the authenticity and integrity of digital content. Existing versatile watermarking approaches suffer from trade-offs between tamper localization precision and visual quality. Constrained by the limited flexibility of previous framework, their localized watermark must remain fixed across all images. Under AIGC-editing, their copyright extraction accuracy is also unsatisfactory. To address these challenges, we propose \textbf{OmniGuard}, a novel augmented versatile watermarking approach that integrates proactive embedding with passive, blind extraction for robust copyright protection and tamper localization. OmniGuard employs a hybrid forensic framework that enables flexible localization watermark selection and introduces a degradation-aware tamper extraction network for precise localization under challenging conditions. Additionally, a lightweight AIGC-editing simulation layer is designed to enhance robustness across global and local editing. Extensive experiments show that OmniGuard achieves superior fidelity, robustness, and flexibility. Compared to the recent state-of-the-art approach EditGuard, our method outperforms it by \textbf{4.25dB} in PSNR of the container image, \textbf{20.7\%} in F1-Score under noisy conditions, and \textbf{14.8\%} in average bit accuracy. \renewcommand\thefootnote{\relax} \footnote{*: Equal contribution, \Letter: Corresponding author.}
\end{abstract}
    
\section{Introduction}
\label{sec:intro}

With the support of massive datasets and the advancement of large-scale model technologies, generative AI has empowered many industries, creating numerous text-to-image models~\cite{dalle3, podellsdxl, rombach2021highresolution} and image-editing algorithms~\cite{brooks2022instructpix2pix, zhang2024magicbrush, liu2024magicquill}. This powerful generative and editing capability is a double-edged sword and has introduced significant risks to the authenticity and integrity of information. For example, illegal infringers may steal images generated by others online and claim ownership, making it increasingly challenging to protect intellectual copyrights. Additionally, fraudsters might use AI tools to alter online images to generate misleading content, presenting new challenges for court forensics.

\begin{figure}[t]
\centering
\includegraphics[width=\linewidth]{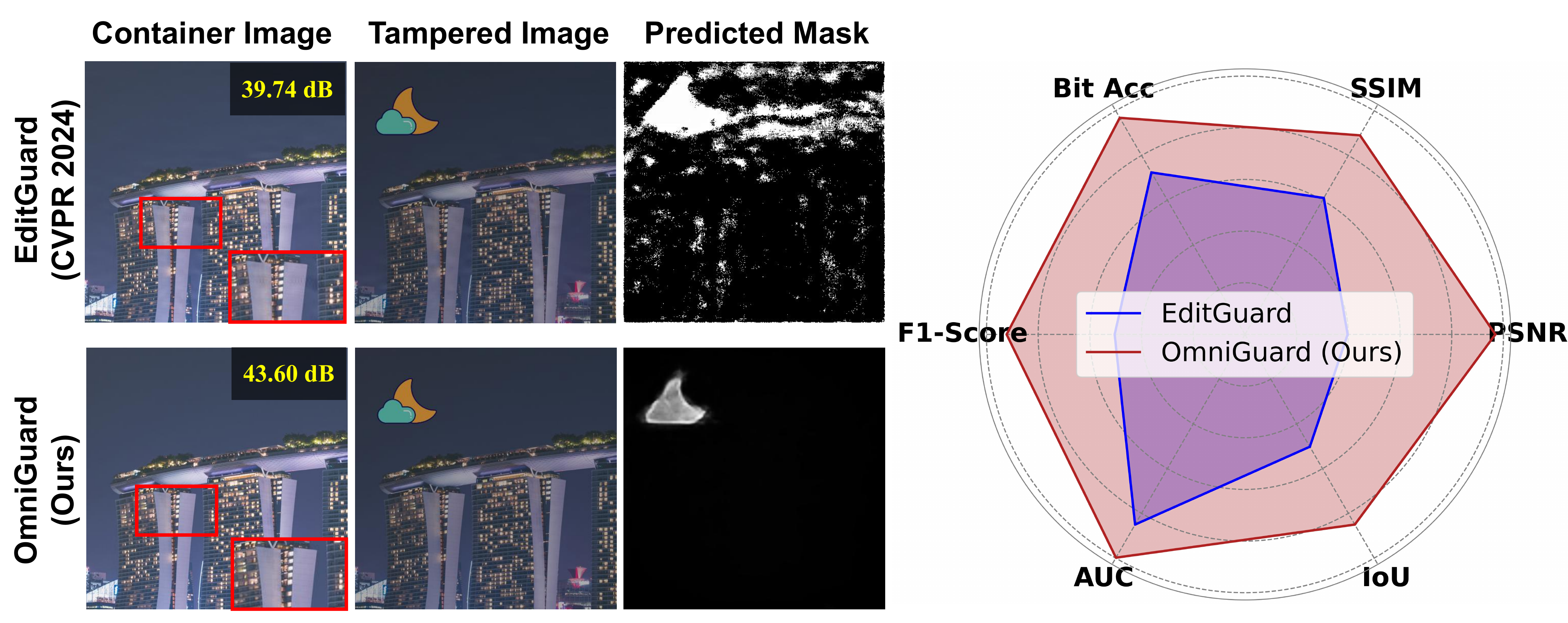}
\vspace{-15pt}
\caption{\textbf{Comparison between OmniGuard and typical state-of-the-art versatile deep watermarking method~\cite{zhang2024editguard}.} Here, we paste a moon onto the container image (tampering) and reduce the overall brightness (degradation). Under AIGC-Editing or simple tampering, our method significantly outperforms EditGuard \cite{zhang2024editguard} across multiple metrics, including container image fidelity (PSNR, SSIM), localization accuracy in degraded conditions (F1-Score, AUC, IoU), and copyright recovery precision (Bit Acc).}
\label{fig:teaser}
\vspace{-10pt}
\end{figure}

As a classic and widely adopted technique, watermarking has drawn growing attention due to its critical role in protecting image copyrights, detecting unauthorized usage, and tracking tampering. Robust watermarking, for instance, focuses on resilience, aiming to keep the embedded watermark intact and detectable even after significant distortions, such as JPEG compression~\cite{jia2021mbrs} or screen shooting~\cite{fang2022pimog}. In contrast, fragile watermarking prioritizes sensitivity, ensuring that the embedded watermark responds to even minor or specific alterations~\cite{liew2013tamper,kamili2020dwfcat,nr2022fragile,lin2023fragile}. Recently, with the development of deep learning, deep watermarking has shown significant advantages in terms of capacity~\cite{xu2022robust}, robustness~\cite{hu2025robust}, and fidelity~\cite{jing2021hinet}. It holds great potential to withstand numerous AIGC editing methods while enabling pixel-wise tamper localization. Notably, some watermarking methods address tamper localization and copyright protection tasks simultaneously~\cite{hurrah2019dual,kamili2020dwfcat, sander2024watermarklocalizedmessages}. For instance, AudioSeal~\cite{san2024proactive} employed a zero-bit watermarking strategy to achieve sample-level audio tamper localization with robust bit embedding. SepMark~\cite{wu2023sepmark} introduced the first deep separable watermark for robust copyright protection and deepfake detection. EditGuard~\cite{zhang2024editguard, zhang2024v2amark} leverages the locality of image-in-image steganography~\cite{ye2024pprsteg} to design a serial encoding and parallel decoding watermarking framework.

However, existing versatile deep image watermarking methods still face several challenges. \textbf{1) Fidelity:} Versatile image watermarking faces an inevitable trade-off between the accuracy of tamper localization and the fidelity of the watermarked image. This is because current methods like~\cite{zhang2024editguard, zhang2024v2amark, zhao2024proactive} rely on residual subtraction between pre-defined localization watermarks and reconstructed ones to generate the mask. Therefore, to ensure satisfactory localization accuracy, the fidelity of the watermarked image has to be sacrificed. \textbf{2) Flexibility:} Since the pre-defined localization watermark must be known at the decoding stage to extract the mask, the localized watermark tends to remain fixed across all images, significantly limiting the flexibility of information embedding. \textbf{3)  Robustness:} Existing versatile watermarking methods exhibit clear weaknesses in robustness. The localized watermark recovery often fails under severe degradations, such as brightness adjustments or heavy image compression. Additionally, the copyright watermark can be erased by global AIGC editing algorithms, further compromising its effectiveness.

To solve the above-mentioned issues, we propose a comprehensively enhanced versatile watermarking approach, dubbed \textbf{OmniGuard}. \textbf{Firstly}, we propose a hybrid forensic approach that combines proactive watermark embedding with passive network-based blind extraction. It can output a tampered mask from the reconstructed localized watermark without prior knowledge of the pre-added one, decoupling the encoding and decoding end. Meanwhile, we train this mask extractor to achieve precise extraction even under inaccurate reconstruction, allowing the watermarking network to focus on enhancing the fidelity of the watermarked image. \textbf{Furthermore}, we explore the selection patterns for localized watermarks and design an adaptive watermark transform, enabling the network to embed localized information in a content-aware manner. \textbf{Finally}, we propose a lightweight AIGC editing simulation layer that enhances the network's copyright extraction accuracy, ensuring robust performance under global editing such as Instructpix2pix~\cite{brooks2022instructpix2pix} and local inpainting like Stable Diffusion Inpaint~\cite{rombach2021highresolution}. Fig.~\ref{fig:teaser} presents our significant improvement on the SOTA versatile watermarking~\cite{zhang2024editguard}. In a nutshell, our contributions are summarized as follows.

\vspace{1pt}
\noindent \ding{113}~We propose a hybrid manipulation localization and robust copyright protection framework OmniGuard, which includes a proactive dual watermark network and a passive extractor, improving existing versatile watermarking.

\vspace{1pt}
\noindent \ding{113}~To integrate proactive watermark with passive extraction into a unified framework, a deep degradation-aware tamper extractor is proposed, which fuses the artifacts from the reconstructed localized watermark with the tampered image, achieving higher accuracy under severe degradations.

\vspace{1pt}
\noindent \ding{113}~To enhance the proactive dual watermark network, an adaptive watermark transform and a lightweight AIGC-edit simulator are designed to respectively enhance the container image fidelity and copyright extraction precision.

\vspace{1pt}
\noindent \ding{113}~Experiments verify that our OmniGuard improves fidelity, flexibility, and robustness compared to existing versatile watermarking. It also surpasses passive localization methods and robust watermarking comprehensively.

\section{Related Works}
\label{sec:related}

\subsection{Image Tamper Detection and Localization}
\textbf{Passive Methods:} Passive image tamper detection and localization networks primarily focus on identifying anomalous regions, such as artifacts, noise, and resolution inconsistencies~\cite{wu2022robust, salloum2018image, islam2020doa,kwon2021cat, wu2019mantra, hu2023draw,xu2024fakeshield, yan2023ucf, kong2022detect}. For instance, MVSS-Net~\citep{dong2022mvss} used multi-scale supervision and multi-view feature learning to capture noise and boundary discrepancies. TruFor~\cite{guillaro2023trufor} employed a noise-sensitive fingerprint and combined high-level and low-level trace extraction via a transformer-driven fusion mechanism. HiFi-Net~\citep{guo2023hierarchical} combined multi-branch feature extraction with specialized localization modules, effectively detecting modifications from CNN-generated or edited images. Meanwhile, IML-ViT~\citep{ma2023iml} incorporated Swin-ViT into its framework with an FPN structure and edge loss constraints to enhance accuracy. Inspired by diffusion models, DiffForensics~\citep{yu2024diffforensics} adopted diffusion denoising pre-training to improve the detection of fine-grained details in tampered images. 

\noindent \textbf{Proactive Schemes:} Proactive tamper detection and localization~\cite{zhao2024proactive, Asnani_2022_CVPR, asnani2024proactive, Asnani_2024_CVPR, ying2022rwn} involves embedding information into the original image before any alterations, which aids the network in detecting manipulations or performing attributions. Although traditional fragile watermarking methods~\cite{cheng2012refining,liew2013tamper,hurrah2019dual,kamili2020dwfcat,nr2022fragile,lin2023fragile} can perform block-wise tamper localization, but their precision and flexibility remain limited. Recently, MaLP~\cite{asnani2023malp} learned the template by leveraging local and global-level features estimated by a two-branch architecture and detect the tampered region from the recovered template. Imuge~\cite{ying2023learning, ying2021image} adopted the self-embedding mechanism and an efficient attack layer to realize tamper localization and self-recovery. Draw~\cite{hu2023draw} embedded watermarks in RAW images to enhance passive tamper localization networks' resilience to lossy operations like JPEG compression, blurring, and re-scaling. EditGuard~\cite{zhang2024editguard} and V$^{2}$AMark~\cite{zhang2024v2amark} used invertible networks to embed localization and copyright watermarks, achieving pixel-wise localization and copyright protection. While these proactive mechanisms have shown promising results, achieving a balance between fidelity, robustness, and localization accuracy remains an open challenge.

\subsection{Deep Image Watermarking}
Image watermarking is a widely accepted and classic technique for copyright protection. Traditional methods embedded secret messages in spatial or adaptive domains~\cite{lsb}, typically placing the data in less significant bits~\cite{stc} or inconspicuous regions, which restricts the capacity for hidden information. Recently, deep image watermarking has gained attention. For example, HiDDeN~\cite{zhu2018hidden} introduced a deep encoder-decoder network to conceal and retrieve bitstream data. Furthermore, flow-based models~\cite{fang2023flow, ma2022towards} were introduced benefited from its inherent information-hiding capability, further enhancing the fidelity of watermarked images. Additionally, various distortion layers like differentiable JPEG and screen-shooting~\cite{liu2019novel, ahmadi2020redmark, wu2023sepmark, fang2022pimog} have been created to boost the robustness. 
SSL~\cite{fernandez2022watermarking} watermarked images within self-supervised latent spaces, enabling enhanced security and resilience via the learned image representations. TrustMark~\cite{bui2023trustmark} presented a universal watermarking technique designed for arbitrary-resolution images. However, these methods cannot effectively support AIGC editing algorithms. Recently, Robust-wide~\cite{hu2025robust} proposed a partial instruction-driven denoising sampling guidance module to encourage robust watermarking against instruction-driven image editing. VINE~\cite{lu2024robust} introduced SDXL-Turbo~\cite{sauer2025adversarial} and a decoder, achieving great bit recovery performance. However, these methods require adding the multi-step iterative diffusion process into network training, significantly reducing training speed and adding computational overhead.

\section{Methodology}
\begin{figure}[t!]
    \centering
    \includegraphics[width=1\linewidth]{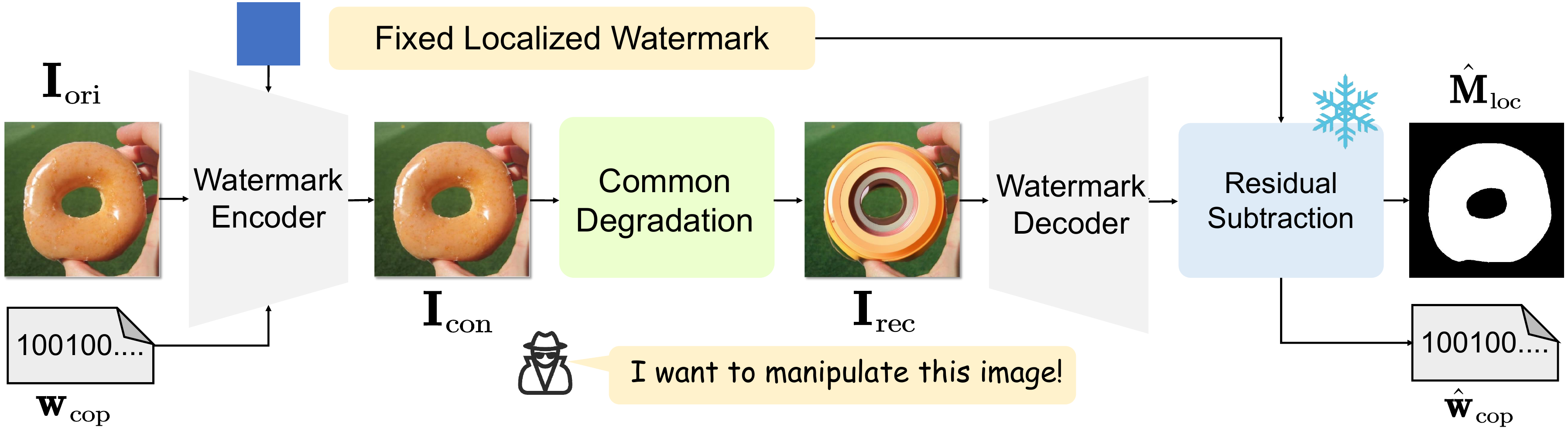}
    \vspace{-15pt}
    \caption{\textbf{Review of prior deep versatile watermarking~\cite{zhang2024editguard, zhao2024proactive}.} They used a fixed localized watermark and produced masks via residual subtraction between the recovered and added ones.}
    \label{fig:review}
    \vspace{-10pt}
\end{figure}

\begin{figure*}[t!]
    \centering
    \includegraphics[width=1\linewidth]{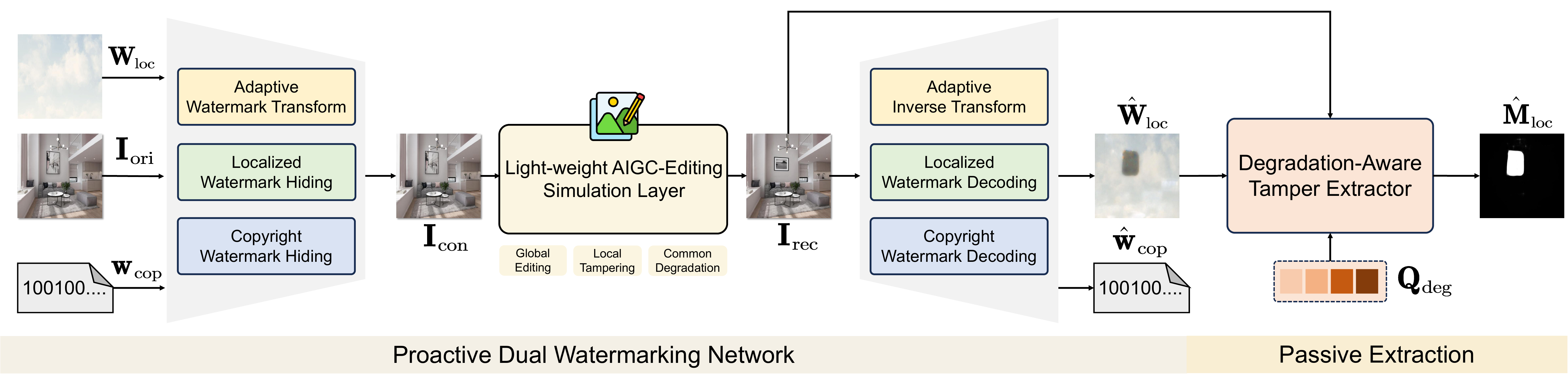}
    \vspace{-15pt}
    \caption{\textbf{High-Level Framework of the proposed OmniGuard.} Based on the existing versatile watermarking framework, we can more flexibly select and embed localized watermarks to improve container image fidelity. Additionally, a deep tamper extractor is introduced to robustly extract tamper masks. The copyright watermark extraction can also resist interference from AIGC-Editing.}
    \label{fig:framework}
    \vspace{-5pt}
\end{figure*}

\subsection{Review of Versatile Deep Image Watermarking}
We first review existing approaches in deep versatile watermarking~\cite{zhang2024editguard, zhang2024v2amark, zhao2024proactive, asnani2024proactive, wu2023sepmark}. Represented by EditGuard~\cite{zhang2024editguard}, current versatile watermarks use a serial encoding and parallel decoding framework, allowing localized and copyright watermarks to coexist within a single image without interfering. Specifically, as shown in Fig.~\ref{fig:review}, given a 2D localized watermark \( \mathbf{W}_{\text{loc}}\)~$\in$~\(\mathbb{R}^{H \times W \times 3}\) and a 1D copyright watermark \(\mathbf{w}_{\text{cop}}\)~$\in$~\(\{0, 1\}^L\), the original image \( \mathbf{I}_{\text{ori}}\)~$\in$~\(\mathbb{R}^{H \times W \times 3}\) is fed to an image-bit united embedding network to produce the container image \( \mathbf{I}_{\text{con}} \). Subsequently, if the received image \( \mathbf{I}_{\text{rec}} \) has undergone local editing and channel degradation, we can input \( \mathbf{I}_{\text{rec}} \) into the image-bit decoding network to retrieve the dual watermarks $\hat{\mathbf{W}}_{\text{loc}}$, $\hat{\mathbf{w}}_{\text{cop}}$. The AIGC editing and degraded pipeline can be modeled as follows.
{\setlength\abovedisplayskip{0.2cm}
\setlength\belowdisplayskip{0.2cm}
\begin{equation}
\mathbf{I}_{\text{rec}}=\mathcal{D}_{\text{deg}}(\mathbf{I}_{\text{con}} \odot (\mathbf{1} - \mathbf{M}_{\text{gt}}) + \mathcal{E}_{\text{edit}}(\mathbf{I}_{\text{con}}) \odot \mathbf{M}_{\text{gt}}), 
    \label{eq1}
\end{equation}}where $\mathcal{E}_{\text{edit}}(\cdot)$, $\mathcal{D}_{\text{deg}}(\cdot)$ and $\mathbf{M}_{\text{gt}}$ respectively denote the editing function, degradation function and tampered mask. Note that these versatile methods~\cite{zhang2024editguard, zhang2024v2amark, zhao2024proactive} only consider local edits, \emph{i.e.}, \( \mathbf{M}_{\text{gt}}\)~\(\neq\)~\( \mathbf{0} \). The recovered copyright $\hat{\mathbf{w}}_{\text{cop}}$ is expected to be generally consistent with $\mathbf{w}_{\text{cop}}$. Finally, we compare $\hat{\mathbf{W}}_{\text{loc}}$ and $\mathbf{W}_{\text{loc}}$ via the residual subtraction to compute the tampered area $\hat{\mathbf{M}}_{\text{loc}}$~$\in$~$\mathbb{R}^{H \times W}$.
{\setlength\abovedisplayskip{0.2cm}
\setlength\belowdisplayskip{0.2cm}
\begin{equation}
\hat{\mathbf{M}}_{\text{loc}}=\Theta_\tau (|\hat{\mathbf{W}}_{\text{loc}}-\mathbf{W}_{\text{loc}}|),
\end{equation}}where $\Theta_\tau(z)$$=$$1$ ($z \geq \tau)$. $|\cdot|$ is an absolute value operation. However, we summarize that this pixel-wise residual subtraction has three major drawbacks.

\vspace{1pt}
\noindent \ding{113}~The recovery of \(\hat{\mathbf{W}}_{\text{loc}}\) and the generation of $\mathbf{I}_{\text{con}}$ are mutually constraining. Considering that detection accuracy is highly dependent on the restoration precision of \(\hat{\mathbf{W}}_{\text{loc}}\), the quality of $\mathbf{I}_{\text{con}}$ has to be significantly compromised.

\vspace{1pt}
\noindent \ding{113}~The extraction of \(\hat{\mathbf{M}}_{\text{loc}}\) requires knowledge of \(\mathbf{W}_{\text{loc}}\), making it non-blind. Therefore, to facilitate collaboration between the decoding and encoding end, a simple approach is to select a fixed \(\mathbf{W}_{\text{loc}}\). However, a fixed localized watermark cannot achieve optimal fidelity for all images.

\vspace{1pt}
\noindent \ding{113}~If the unaltered regions of \( \hat{\mathbf{W}}_{\mathrm{loc}} \) contain artifacts due to severe degradation, these areas may also be incorrectly treated as tampered, posing a threat to detection robustness.

\begin{table}[t!]
\centering
\resizebox{0.95\linewidth}{!}{
\begin{tabular}{lcc}
\toprule[1.5pt]
Method & EditGuard & \textbf{OmniGuard} \\ \hline
Localized Watermark  & Fixed & \textbf{Adaptive}   \\
Supported Degradation & Common & \textbf{Common \& AIGC-Edit} \\ 
Mask Extraction & Fragile & \textbf{Robust} \\ \bottomrule[1.5pt]
\end{tabular}}
\vspace{-5pt}
\caption{\textbf{High-level comparison between two approaches.}}
\vspace{-10pt}
\label{feature}
\end{table}

\subsection{Overall Framework of OmniGuard}
To address the above issues, we propose a hybrid forensic framework, OmniGuard, which combines a proactive dual watermarking network with a passive, deep, degradation-aware tamper extractor (Fig.~\ref{fig:framework}). This framework leverages the strong generalization capabilities of proactive watermarking to overcome the accuracy limitations of passive extraction networks while using the passive network to further enhance localization robustness. Additionally, OmniGuard improves container image fidelity through a carefully designed localization watermark and strengthens copyright extraction robustness by effectively simulating AIGC global and local edits within the degradation layer.

\begin{figure*}[t!]
    \centering
    \includegraphics[width=1\linewidth]{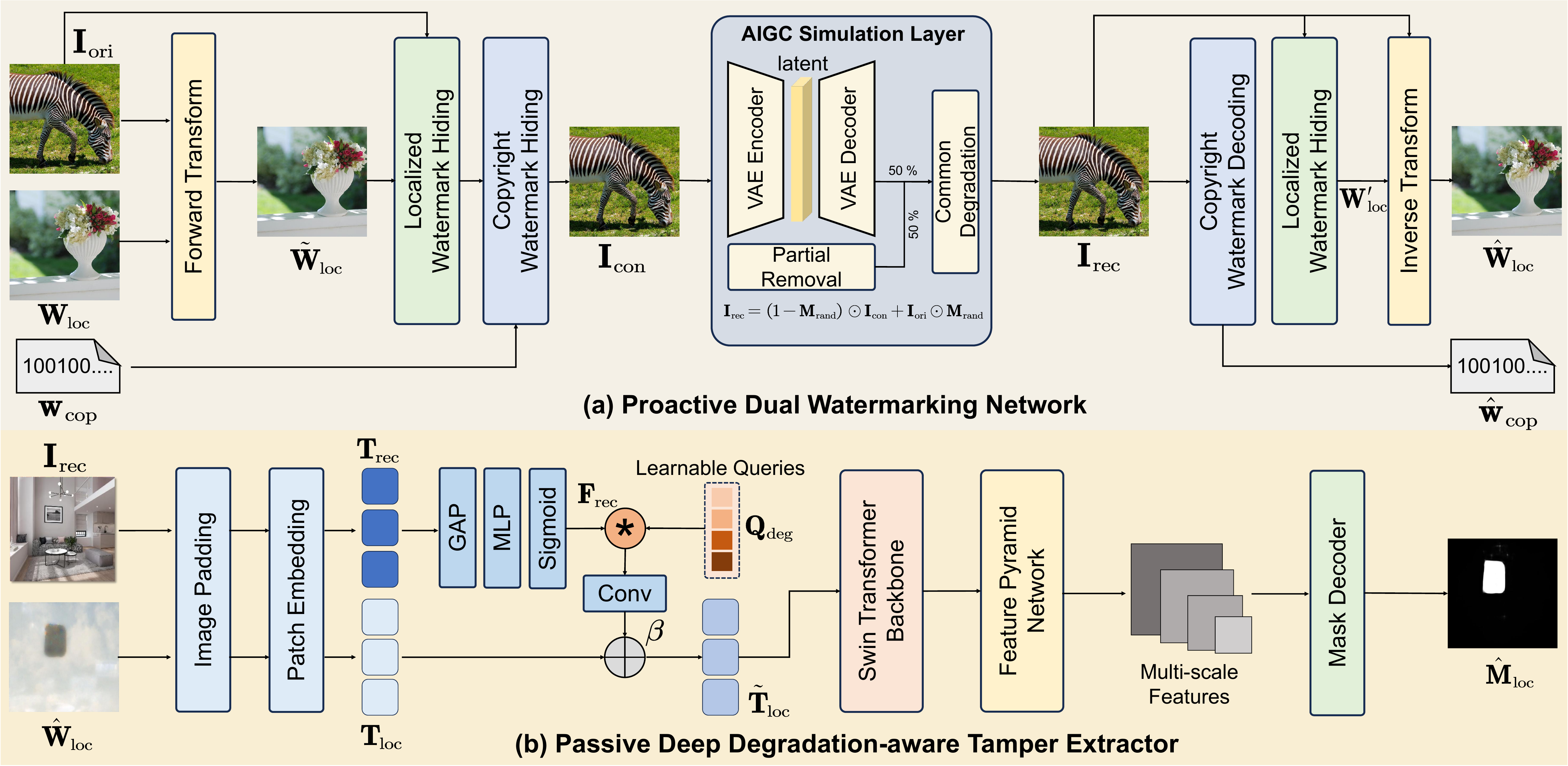}
    \vspace{-15pt}
    \caption{\textbf{Detail of our proposed OmniGuard.} Our proactive dual watermarking network utilizes forward and backward transform to filter high-frequency information from the localized watermark. The passive degradation-aware tamper extractor employs a window-based transformer and degradation querying techniques to predict the mask $\hat{\mathbf{M}}_{\text{loc}}$ from $\hat{\mathbf{W}}_{\text{loc}}$ predicted by the trained watermarking network and $\mathbf{I}_{\text{rec}}$. The AIGC-Editing simulator uses partial removal and VAE compression as surrogate attacks for global edits and local tampering.}
    \label{fig:detail}
    \vspace{-5pt}
\end{figure*}

Specifically, as shown in Fig.~\ref{fig:framework} and Fig.~\ref{fig:detail}(a), we first use an adaptive forward transform (Sec.~\ref{ada_watermark}) to embed the original image $\mathbf{I}_{\text{ori}}$ into the localized watermark $\mathbf{W}_{\text{loc}}$, allowing the transformed watermark $\tilde{\mathbf{W}}_{\mathrm{loc}}$ to acquire some content-aware and reasonable texture, which has been shown to facilitate effective hiding. Then, the localized tag $\tilde{\mathbf{W}}_{\text{loc}}$ and a copyright watermark $\mathbf{w}_{\text{cop}}$ are embedded into $\mathbf{I}_{\text{ori}}$ to create the container image $\mathbf{I}_{\text{con}}$. Similarly, we extract \(\hat{\mathbf{w}}_{\text{cop}}\) from \(\mathbf{I}_{\text{rec}}\) and obtain the artifact map \(\hat{\mathbf{W}}_{\text{loc}}\) via the localized watermark decoding module and an inverse transform (Sec.~\ref{ada_watermark}). Here the localized watermark hiding and decoding networks follow the structure of invertible networks in~\cite{zhang2024editguard}. The structure of copyright watermark hiding and decoding network are inspired by~\cite{bui2023trustmark}. Finally, both the artifact map $\hat{\mathbf{W}}_{\text{loc}}$ and the tampered image $\mathbf{I}_{\text{rec}}$ are fed to the deep tamper extractor to produce the tampered mask $\hat{\mathbf{M}}_{\text{loc}}$.

\textbf{The major improvements of OmniGuard focus on two key points.} \textbf{First}, we introduce a deep tamper extractor that takes both the tampered image \(\mathbf{I}_{\text{rec}}\) and the artifact map \(\hat{\mathbf{W}}_{\text{loc}}\) to analyze tampering clues. By reframing the reconstruction of \(\hat{\mathbf{W}}_{\text{loc}}\) as a classification task, the extractor only needs to distinguish tampering artifacts from other regions, greatly simplifying training. \textbf{Second}, we expand the range of degradations that EditGuard supports. In copyright decoding, we design a lightweight AIGC simulation layer, enabling OmniGuard to support both global edits and local tampering. For localized watermark decoding, benefiting from the passive extractor, our detection can withstand conditions such as JPEG ($\mathrm{Q}$=50), color jitter, and severe noise. We list the comparison with EditGuard on Tab.~\ref{feature}.

\subsection{Passive Degradation-Aware Tamper Extractor}
\noindent \textbf{Movtivation:} The core design of our deep passive extraction network lies in the introduction of $\mathbf{I}_{\text{rec}}$ and a learnable degradation query $\mathbf{Q}_{\text{deg}}$. Considering that the artifact map $\hat{\mathbf{W}}_{\text{loc}}$ often does not reconstruct well under some severe degradations, we also include the original tampered image $\mathbf{I}_{\text{rec}}$ as an auxiliary input to help the network distinguish between artifacts caused by tampering or insufficient reconstruction. Meanwhile, inspired by Q-Former~\cite{li2023blip}, we learn a degradation representation space $\mathbf{Q}_{\text{deg}}$ via the supervision of the \{$\hat{\mathbf{W}}_{\text{loc}}$, $\mathbf{I}_{\text{rec}}$, $\mathbf{M}_{\text{gt}}$\} pairs and adaptively select the degradation type based on the content of \( \mathbf{I}_{\text{rec}} \).

As shown in Fig.~\ref{fig:detail}(b), motivated by~\cite{ma2023iml}, we use a window-based transformer to better capture long-range dependencies and extract tampering traces, enabling precise mask extraction. Specifically, $\hat{\mathbf{W}}_{\text{loc}}$ and $\mathbf{I}_{\text{rec}}$ will be padded in a fixed resolution and perform patch embedding to transform the two images into two groups of image tokens $\mathbf{T}_{\text{loc}}$, $\mathbf{T}_{\text{rec}}$. To adaptively query the degradation type in $\mathbf{Q}_{\text{deg}}$ that corresponds to the content of the input tampered image $\mathbf{I}_{\text{rec}}$, $\mathbf{I}_{\text{rec}}$ is passed to a global average pooling (GAP) layer, an MLP layer, and a sigmoid operator to align with the degradation query $\mathbf{Q}_{\text{deg}}$, extracting the feature $\mathbf{F}_{\text{rec}}$. Instead of using cross-attention, we use a simple element-wise multiplication and a convolution layer to fuse the feature $\mathbf{F}_{\text{rec}}$ and degradation query $\mathbf{Q}_{\text{deg}}$, forming the enhanced artifact map token $\tilde{\mathbf{T}}_{\text{loc}}$. The process is formulated as follows.
{\setlength\abovedisplayskip{0.15cm}
\setlength\belowdisplayskip{0.15cm}
\begin{equation}
\begin{gathered}
\tilde{\mathbf{T}}_{\text{loc}}=\mathbf{T}_{\text{loc}}+ \beta\cdot\operatorname{Conv}(\mathbf{F}_{\text{rec}}\circledast \mathbf{Q}_{\text{deg}}), \\
with~~~ \mathbf{F}_{\text{rec}}=\operatorname{Sigmoid}\left(\operatorname{MLP}\left(\operatorname{GAP}\left(\mathbf{T}_{\text{rec}}\right)\right)\right),
\end{gathered}
\end{equation}}where $\beta$ is a learnable trade-off parameter, $\circledast$ denotes element-wise multiplication. Furthermore, we feed the enhanced token $\tilde{\mathbf{T}}_{\text{rec}}$ into a Swin-ViT backbone~\cite{li2022mvitv2} for visual feature extraction and use a feature pyramid network to obtain multi-scale features. These multi-scale features are then processed by a mask decoder, composed of several linear layers, to generate the final mask $\hat{\mathbf{M}}_{\text{loc}}$. The entire process of our mask extraction is expressed as follows.
{\setlength\abovedisplayskip{0.15cm}
\setlength\belowdisplayskip{0.15cm}
\begin{equation}
    \hat{\mathbf{M}}_{\text{loc}} = \operatorname{TamperExtracor}(\hat{\mathbf{W}}_{\text{loc}}~|~\mathbf{I}_{\text{rec}}, \mathbf{Q}_{\text{deg}})
\end{equation}}

\subsection{Adaptive Localized Watermark Transform}
\label{ada_watermark}
\textbf{Motivation:} Previous methods, limited by the need for residual subtraction to generate the mask, often rely on fixed, simple-textured localized watermarks (\emph{i.e.}~a plain blue image) to ensure sufficient precision in the reconstructed watermark. However, we observe that embedding a natural image (\emph{i.e.}~a sky with white clouds image in Fig.~\ref{fig:framework}) is often less noticeable, resulting in a more natural pattern of artifacts. Adopting our robust passive extraction method, we can use a more complex localized image that better complements the original content. Additionally, one-step steganographic embedding can produce content-aware noise while filtering out high-frequency details from the watermark, making it more suitable for information hiding.

Specifically, as plotted in Fig.~\ref{fig:detail}(a), our forward transform and inverse transform form a pair of reversible additive affine transformations. We first hide the $\mathbf{I}_{\text{ori}}$ into $\mathbf{W}_{\text{loc}}$ to produce an more adaptive localized watermark $\tilde{\mathbf{W}}_{\text{loc}}$. 
{\setlength\abovedisplayskip{0.15cm}
\setlength\belowdisplayskip{0.15cm}
\begin{align}
& \tilde{\mathbf{W}}_{\text{loc}} = \mathbf{W}_{\text{loc}} + \phi_1( \mathbf{I}_{\text{ori}}), \nonumber \\
& \mathbf{H}_{\text{med}} = \mathbf{I}_{\text{ori}} \otimes \operatorname{Exp}(\phi_2(\tilde{\mathbf{W}}_{\text{loc}})) +\phi_3(\tilde{\mathbf{W}}_{\text{loc}}), 
\label{eq2}
\end{align}}where $\{\phi_i\}_{i=1}^3$ respectively denote a simple convolution layer to perform affine transformation. $\mathbf{H}_{\text{med}}$ denotes the intermediate dismissed high-frequency information. Similarly, in the inverse transform, we use $\mathbf{I}_{\text{rec}}$ to approximate $\mathbf{I}_{\text{ori}}$, and perform further computation based on the result of the localized watermark decoding $\mathbf{W}^{\prime}_{\text{loc}}$ as follows.
{\setlength\abovedisplayskip{0.15cm}
\setlength\belowdisplayskip{0.15cm}
\begin{align}
    & \hat{\mathbf{H}}_{\text{med}} = (\mathbf{I}_{\text{rec}}-\phi_3 (\mathbf{W}^{\prime}_{\text{loc}})) \otimes \operatorname{Exp}( -\phi_2(\mathbf{W}^{\prime}_{\text{loc}})), \nonumber \\
    & \hat{\mathbf{W}}_{\text{loc}} = \mathbf{W}^{\prime}_{\text{loc}} - \phi_1(\mathbf{I}_{\text{rec}}).
\label{eq3}
\end{align}}Through a one-step forward and inverse transform, the pre-selected watermark $\mathbf{W}_{\text{loc}}$ will undergo slight adaptive adjustments and be conducive to the fidelity of the container image without compromising localization accuracy.

\begin{table*}[t!]
    \renewcommand{\arraystretch}{1.1}
    \centering
        \resizebox{0.9\linewidth}{!}{
    \begin{tabular}{@{}lccccccccccccccc@{}}
        \toprule[1.5pt]
        \multirow{3}{*}{Method} & \multicolumn{4}{c}{Stable Diffusion Inpaint} & \multicolumn{4}{c}{Controlnet Inpaint} & \multicolumn{4}{c}{Splicing} \\
        \cmidrule(lr){2-13}
        & \multicolumn{2}{c}{Clean} & \multicolumn{2}{c}{Degraded} & \multicolumn{2}{c}{Clean} & \multicolumn{2}{c}{Degraded} & \multicolumn{2}{c}{Clean} & \multicolumn{2}{c}{Degraded} \\
        \cmidrule(lr){2-3} \cmidrule(lr){4-5} \cmidrule(lr){6-7} \cmidrule(lr){8-9} \cmidrule(lr){10-11} \cmidrule(lr){12-13}
        & F1 & AUC & F1 & AUC & F1 & AUC & F1 & AUC & F1 & AUC & F1 & AUC \\
        \midrule
        MVSS-Net~\cite{dong2022mvss} & 0.183 & 0.735 & 0.181 & 0.741 & 0.226 & 0.765 &  0.208 & 0.750 & 0.355 & 0.825 & 0.352 & 0.823 \\
        CAT-Net~\cite{kwon2021cat} & 0.153 & 0.714 & 0.119 & 0.706 & 0.172 & 0.728 & 0.130 & 0.710 & 0.198 & 0.735 & 0.196 & 0.734 \\
        PSCC-Net~\cite{liu2022pscc} & 0.098 & 0.630 & 0.041 & 0.554 & 0.086 & 0.624 & 0.086 & 0.627 & 0.083 & 0.618 & 0.082 & 0.602  \\
        HiFi-Net~\cite{guo2023hierarchical} & 0.188 & 0.534 & 0.193 & 0.549 & 0.190 & 0.546 & 0.190 & 0.547 & 0.189 & 0.553 & 0.189  & 0.553 \\
        IML-ViT~\cite{ma2023iml} & 0.099 & 0.636 & 0.107 & 0.643 & 0.111 & 0.643 & 0.111 & 0.650 & 0.352 & 0.775 & 0.348 & 0.751 \\
        EditGuard~\cite{zhang2024editguard}   &\underline{\textcolor{blue}{0.951}} &\underline{\textcolor{blue}{0.971}} &\underline{\textcolor{blue}{0.740}} &\underline{\textcolor{blue}{0.925}} &\textcolor{red}{\textbf{0.958}} &\underline{\textcolor{blue}{0.987}} &\underline{\textcolor{blue}{0.752}} &\underline{\textcolor{blue}{0.921}} &\textcolor{red}{\textbf{0.928}} &\underline{\textcolor{blue}{0.990}} &\underline{\textcolor{blue}{0.758}} &\underline{\textcolor{blue}{0.933}} \\
        OmniGuard (Ours) &\textcolor{red}{\textbf{0.961}} &\textcolor{red}{\textbf{0.999}} &\textcolor{red}{\textbf{0.947}} &\textcolor{red}{\textbf{0.998}} &\underline{\textcolor{blue}{0.953}} &\textcolor{red}{\textbf{0.998}} &\textcolor{red}{\textbf{0.952}} &\textcolor{red}{\textbf{0.999}} &\underline{\textcolor{blue}{0.913}} &\textcolor{red}{\textbf{0.993}} &\textcolor{red}{\textbf{0.895}} &\textcolor{red}{\textbf{0.991}} \\
        \bottomrule[1.5pt]
    \end{tabular}}
    \vspace{-5pt}
    \caption{\textbf{Localization performance of the proposed OmniGuard and other SOTA proactive or passive manipulation localization methods.} ``Clean'' and ``Degraded'' denote detection under the clean condition, and under the condition of randomly selecting JPEG, color jitter, Gaussian noise, and salt-pepper noise. The best results are highlighted in \textcolor{red}{\textbf{red}}, and the second one is \underline{\textcolor{blue}{blue}}.}
    \vspace{-5pt}
    \label{localization}
\end{table*}

\subsection{Light-Weight AIGC-Editing Simulator}
\textbf{Motivation:} To enhance the copyright watermark’s resilience to AIGC edits, we introduce a lightweight AIGC-editing simulation layer. For global edits, prior methods~\cite{hu2025robust} often incorporate the iterative diffusion denoising process into training, which significantly increases memory consumption and requires substantial image-text pairs to support text-driven edits. However, we observe that the primary source of watermark disruption after editing stems from VAE compression rather than the denoising process itself. Therefore, integrating VAE into the degradation layer can replicate the effects of actual AIGC editing while reducing training time to $\textbf{1/24}$ of Robust-wide~\cite{hu2025robust}. Additionally, previous methods~\cite{zhang2024editguard} typically overlooked local edits during training, relying on the watermark’s global hiding capabilities to mitigate local tampering. However, new content added to local regions introduces interference with watermark decoding. Thus, we incorporate a simple partial watermark removal operator to address it more precisely.

As plotted in Fig.~\ref{fig:detail}(a), the container image $\mathbf{I}_{\text{con}}$ has a 50\% chance of being fed to the Stable Diffusion VAE for encoding and decoding to simulate global editing and a 50\% chance of undergoing a partial removal operation as surrogate attacks. Instead of a straight-through estimator, all the gradients of these operators can be propagated back. Then, we also introduce some common degradations after the AIGC editing. The processes are as follows.
{\setlength\abovedisplayskip{0.15cm}
\setlength\belowdisplayskip{0.15cm}
\begin{equation}
    \mathbf{I}_{\text{rec}} = 
\begin{cases} 
    \mathcal{D}_{\text{com}}(\operatorname{VAE}(\mathbf{I}_{\text{con}})), & \text{if } p>0, \\ 
    \mathcal{D}_{\text{com}}((\mathbf{1} - \mathbf{M}_{\text{rand}})\odot\mathbf{I}_{\text{con}} + \mathbf{I}_{\text{ori}}\odot\mathbf{M}_{\text{rand}}), & \text{else} \\ 
\end{cases}
\end{equation}}where $\operatorname{VAE}$ denotes the variational encoder-decoder of SD-1.5~\cite{rombach2021highresolution} with frozen parameters, $\mathcal{D}_{\text{com}}$ denotes the common degradation function, and $p$~$\sim$~$\mathcal{N}(0, 1)$ denotes the selection probability. $\mathbf{M}_{\text{rand}}$ denotes the  randomly generated mask. Note that we only introduce our lightweight AIGC-Editing simulator in the training of copyright watermark encoding and decoding. When training the dual watermarking network, we only introduce differentiable JPEG~\cite{xu2022robust} to ensure sufficient quality of the container image. 

\subsection{Training Details}
Our training process consists of three stages. First, given the original image and the copyright watermark $\{\mathbf{I}_{\text{ori}}, \mathbf{w}_{\text{cop}}\}$, we train the copyright watermark hiding and decoding network via the binary cross-entropy loss $\ell_{\text{bce}}$ and mean squared error loss $\ell_2$. The loss function is:
{\setlength\abovedisplayskip{0.15cm}
\setlength\belowdisplayskip{0.15cm}
\begin{equation}
    \ell_{\text{cop}} =  \ell_{\text{bce}}(\hat{\mathbf{w}}_{\text{cop}}, \mathbf{w}_{\text{cop}}) + \lambda \cdot \ell_2(\mathbf{I}_{\text{con}}, \mathbf{I}_{\text{ori}}),
\end{equation}}where $\lambda$ is firstly set to $0.05$, and gradually increase to $27.5$. After $150$ epochs, we introduce surrogate attacks like VAE and partial removal. Then, we freeze their weights and jointly train the whole proactive dual watermark network via the image-localized watermark pair $\{\mathbf{I}_{\text{ori}}, \mathbf{W}_{\text{loc}} \}$.
{\setlength\abovedisplayskip{0.15cm}
\setlength\belowdisplayskip{0.15cm}
\begin{align}
    \ell_{\text{loc}} &= \ell_2(\hat{\mathbf{W}}_{\text{loc}}, \mathbf{W}_{\text{loc}}) + \alpha_1 \cdot \ell_2(\mathbf{I}_{\text{con}}, \mathbf{I}_{\text{ori}}) \nonumber \\ 
    &+ \alpha_2\cdot \ell_{\text{GAN}}(\mathbf{I}_{\text{con}}, \mathbf{I}_{\text{ori}}) + \alpha_3\cdot \ell_{\text{LPIPS}}(\mathbf{I}_{\text{con}}, \mathbf{I}_{\text{ori}}),
\end{align}}where $\alpha_1$, $\alpha_2$, and $\alpha_3$ are respectively set to $10$, $10$, and $100$. Since passive extraction has relatively low precision requirements for $\hat{\mathbf{W}}_{\text{loc}}$, we make the network focus more on optimizing the quality of the container image and assign them a higher weight. Additionally, we introduce GAN loss~\cite{fang2022pimog} $\ell_{\text{GAN}}$ and VGG loss $\ell_{\text{LPIPS}}$ to enhance perceptual quality. After training the watermarking part, we train the tamper extractor via the triplet $\{\hat{\mathbf{W}}_{\text{loc}}, \mathbf{I}_{\text{rec}}, \mathbf{M}_{\text{gt}} \}$. 
{\setlength\abovedisplayskip{0.15cm}
\setlength\belowdisplayskip{0.15cm}
\begin{equation}
    \ell_{\text{ext}} =  \ell_{\text{bce}}(\hat{\mathbf{M}}_{\text{loc}}, \mathbf{M}_{\text{gt}}) + \gamma \cdot \ell_{\text{edge}}(\mathbf{M}_{\text{loc}}, \mathbf{M}_{\text{gt}}),
\end{equation}}where $\gamma$ is set to $20$. The edge loss $\ell_{\text{edge}}$ is used to guide the model to focus on important regions~\cite{bui2023trustmark}.

\begin{figure*}[t!]
    \centering
    \includegraphics[width=1\linewidth]{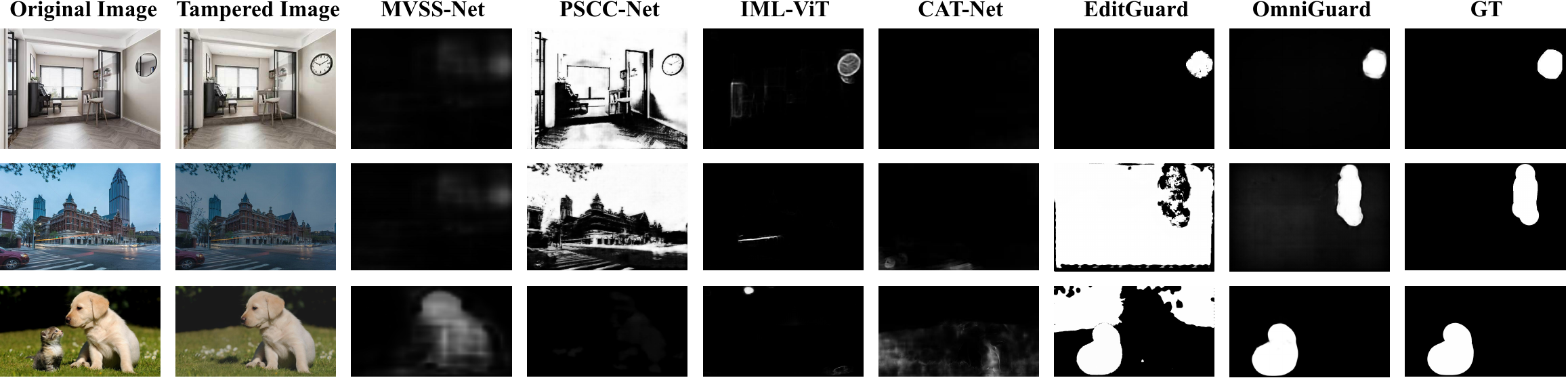}
    \vspace{-20pt}
    \caption{\textbf{Visualized Comparison between our OmniGuard and other methods on the in-the-wild tampered samples.} The tampered image has undergone JPEG compression (1$^{st}$ row), brightness adjustment (2$^{nd}$ row) and contrast adjustment (3$^{rd}$ row).}
    \label{fig:locresult}
    \vspace{-5pt}
\end{figure*}

\section{Experimental Results}

\begin{table*}[t!]
    \renewcommand{\arraystretch}{1.1}
    \centering
    \resizebox{1.\linewidth}{!}{
    \begin{tabular}{@{}lccccccccccc@{}}
        \toprule[1.5pt]
        \multirow{3}{*}{Method} & \multirow{3}{*}{Capacity} & \multirow{3}{*}{PSNR} & \multirow{3}{*}{SSIM} & \multicolumn{6}{c}{Bit Accuracy (\%)} \\
        \cmidrule(lr){5-11}
        & & & & \multicolumn{2}{c}{Global Edit} & \multicolumn{2}{c}{Local Edit} & \multicolumn{3}{c}{Common Degradation} \\
        \cmidrule(lr){5-6} \cmidrule(lr){7-8} \cmidrule(lr){9-11}
        & & & & Instructp2p & SD Inpaint* & SD Inpaint & Random Dropout & JPEG & ColorJitter & Gaussian Noise \\
        \midrule
        PIMoG~\cite{fang2022pimog} &30 bits &36.68 &0.926 &0.695 &0.641 &0.945 &0.947 &0.961 &0.965 &0.756 \\
        SepMark~\cite{wu2023sepmark} &30 bits & 34.71 & 0.890 &0.921 &0.928 &0.978 &0.980 &0.999 &0.964 &\underline{\textcolor{blue}{0.952}} \\
        TrustMark~\cite{bui2023trustmark} & 100 bits &\textcolor{red}{\textbf{43.17}} &\textcolor{red}{\textbf{0.994}} &0.928 &0.939 &0.980 &0.982 &0.998 &0.962 &0.921\\
        EditGuard~\cite{zhang2024editguard} & 64 bits &42.15 &0.986 &0.582 &0.623 &0.982 &0.982 &0.975 & \underline{\textcolor{blue}{0.978}} &0.778 \\
        Robust-Wide~\cite{hu2025robust} & 64 bits &\underline{\textcolor{blue}{42.72}} &0.992 &\textcolor{red}{\textbf{0.990}} &\underline{\textcolor{blue}{0.974}} &\underline{\textcolor{blue}{0.984}} &\underline{\textcolor{blue}{0.986}} &\underline{\textcolor{blue}{0.999}} &0.964 & 0.732 \\
        OmniGuard & 100 bits &42.33 &\underline{\textcolor{blue}{0.992}} &\underline{\textcolor{blue}{0.981}} &\textcolor{red}{\textbf{0.989}} &\textcolor{red}{\textbf{0.996}} &\textcolor{red}{\textbf{0.996}} &\textcolor{red}{\textbf{1.000}} &\textcolor{red}{\textbf{0.980}} &\textcolor{red}{\textbf{0.959}} \\ \midrule[1.5pt]
        EditGuard$^{\dagger}$~\cite{zhang2024editguard} & 64 bits + $\mathbf{W}_{\text{loc}}$ &37.53 &0.936 &0.556 &0.617 &0.981 &0.982 &0.971 &0.977 &0.776 \\
        OmniGuard$^{\dagger}$ &100 bits + $\mathbf{W}_{\text{loc}}$ &41.78 &0.989 &0.979 &0.988 &0.995 &0.994 &1.000 &0.979 &0.959 \\
        \bottomrule[1.5pt]
    \end{tabular}}
    \vspace{-8pt}
    \caption{\textbf{Fidelity and bit recovery accuracy comparison between the proposed OmniGuard and other SOTA watermarking methods.} Note that ``SD Inpaint*'' denotes the regeneration from the image via an inpainting model, while ``SD Inpaint'' ensures that the non-edited regions remain entirely consistent with the original image. $\dagger$ denotes hiding both a localized watermark and copyright watermark. }
    \label{copyright}
    \vspace{-10pt}
\end{table*}

\subsection{Experimental Setup}
We train the copyright watermark network on the MIRFlickR dataset~\cite{huiskes2008mir} with AdamW optimizer at an initial learning rate of $4\times 10^{-6}$ in a batch of $32$ and cosine annealing schedule. Then, we optimize the entire proactive dual watermarking network on the DIV2K~\cite{DIV2K} dataset with Adam optimizer at a learning rate of $1\times 10^{-5}$ and a batch size of $8$. The window-based transformer backbone is ViT-B which is pretrained on ImageNet-1k via MAE~\cite{he2022masked}. The window size and padding size are relatively set to 14 and 1024. The tamper extractor is trained on $20000$ constructed pairs $\{\hat{\mathbf{W}}_{\text{loc}}, \mathbf{I}_{\text{rec}}, \mathbf{M}_{\text{gt}} \}$. Here, the tampering type during training is Stable Diffusion Inpaint, but our method can generalize to almost any type of local modification. We conduct all our experiments on an NVIDIA GTX 3090Ti Server. 

\subsection{Comparision with Localization Methods}
To evaluate the tamper localization precision of our OmniGuard, we compare it with some state-of-the-art passive methods, including PSCC-Net~\cite{liu2022pscc}, MVSS-Net~\cite{dong2022mvss}, CAT-Net~\cite{kwon2021cat}, HiFi-Net~\cite{guo2023hierarchical}, and IML-ViT~\cite{ma2023iml}, as well as the proactive method EditGuard~\cite{zhang2024editguard}. F1-Score and AUC are used as our evaluation metrics. We compare our methods and other competitive methods on $1000$ samples from the testing set of COCO~\cite{lin2014microsoft}. We select three distinctive local editing methods, namely Stable Diffusion Inpaint, ControlNet Inpaint, and image splicing, to realize $\mathcal{E}_{edit}$ in Eq.~\ref{eq1}. The degradation function $\mathcal{D}_{deg}$ is randomly selected from Gaussian noise ($\sigma$=1-10), JPEG ($\mathrm{Q}$=70-85), color jitter in brightness, contrast, and saturation, and salt-and-pepper noise to verify the robustness. 

As reported on Tab.~\ref{localization}, our method achieves the best localization performance on both AIGC and simple tampering, with an F1 score greater than 0.95 and an AUC close to 1. Although EditGuard performs similarly to ours under clean conditions, we observe a significant drop in EditGuard's localization accuracy under noisy conditions, with F1 decreasing by about 0.2. This is because EditGuard relies on pixel-level comparison, making its localization results highly dependent on watermark recovery performance. In contrast, OmniGuard experiences almost no performance reduction under common degradations. Fig.~\ref{fig:locresult} further illustrates the advantages of our method. On some in-the-wild tampered samples, passive methods such as PSCC-Net and IML-ViT, struggle to effectively detect tampering traces. Proactive methods like EditGuard produce inaccurate masks and are sensitive to threshold settings when the tampered images undergo global degradations. In contrast, our OmniGuard can robustly and accurately identify tampered regions. Note that for global edits like Instructp2p, OmniGuard can still robustly retrieve the copyright and detect tampering; however, it tends to locate the entire image as tampering. This aligns with our intended functionality, as our watermarking focuses more on addressing pixel-level tampering rather than semantic modifications.

\begin{figure*}[t!]
    \centering
    \includegraphics[width=0.95\linewidth]{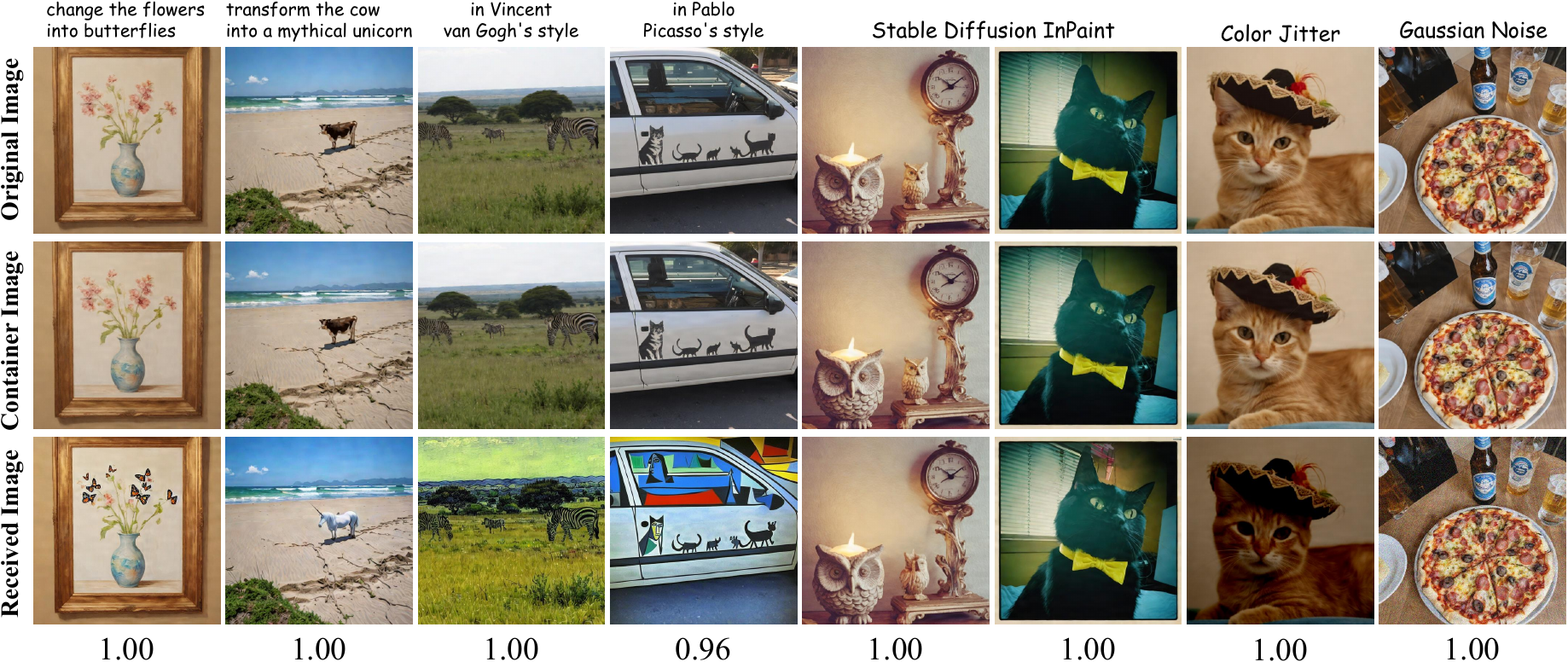}
    \vspace{-8pt}
    \caption{\textbf{Visualization of the original images, container images, and received images (Edited or Degraded) of our OmniGuard.} The recovered bit accuracy is listed below. The first four column is edited via the Instructp2p~\cite{brooks2022instructpix2pix}.}
    \label{fig:bitacc}
    \vspace{-10pt}
\end{figure*}

\subsection{Comparison with Deep Watermarking}
To validate the fidelity and copyright recovery robustness of our method, we compare it with SOTA deep watermarking approaches PIMoG~\cite{fang2022pimog}, SepMark~\cite{wu2023sepmark}, TrustMark~\cite{bui2023trustmark}, EditGuard~\cite{zhang2024editguard} and Robust-Wide~\cite{hu2025robust}. Tab.~\ref{copyright} presents the fidelity of different methods and their bit recovery metrics on three tampering types. Notably, ``SD inpaint*'' denotes regenerating a new image conditioned on a mask. In contrast, local edit strictly follows the operations in Eq.~\ref{eq1}, ensuring that unedited areas remain consistent with the original image. We test all the results on $1000$ $512$~$\times$~$512$ images with paired prompt in the dataset of UltraEdit~\cite{zhao2024ultraeditinstructionbasedfinegrainedimage}. $\sigma$ in Gaussian noise and $\mathrm{Q}$ in JPEG are respectively set to 25 and 70. Color jitter denotes reducing or increasing the brightness by 30\%. More results are shown in \textcolor{blue}{\textbf{supplementary materials}}.

As reported in Tab.~\ref{copyright}, our method achieves the best bit recovery performance across most degradation with satisfactory visual fidelity (PSNR $>$ 42dB). With both the copyright and localized watermark, our method outperforms EditGuard by \textbf{4.25} dB in PSNR and presents a substantial improvement in bit accuracy. On Instructp2p, our bit accuracy is only 0.009 lower than Robust-Wide, which is specifically trained for AIGC Editing. Furthermore, benefiting from our AIGC simulator, the iteration time per sample during training is \textbf{0.16s}, compared to \textbf{3.675s} for Robust-Wide. Moreover, since we do not need to incorporate the diffusion denoising process, we can even train our model on an 11GB single NVIDIA 1080Ti. As plotted in Fig.~\ref{fig:bitacc}, even when the overall style of the image is modified, our OmniGuard can still accurately retrieve the copyright. It effectively manages global and local edits, as well as conventional degradations.

\begin{table}[t!]
\centering
\resizebox{1\linewidth}{!}{
\begin{tabular}{lcccc}
\toprule[1.5pt]
Case & Localized Watermark       & PSNR (dB) & F1 & AUC \\ \hline
(a)  & Fixed Pure Image          &39.64 &\textcolor{red}{\textbf{0.975}}  &\textcolor{red}{\textbf{0.999}}    \\ 
(b)  & Fixed Natural Image       &\underline{\textcolor{blue}{41.09}} &0.957  &0.993 \\
(c)  & Adaptive Natural Image    &40.28 &0.885  &0.975     \\
(d)  & Fixed Natural Image + A.T. (Ours) & \textcolor{red}{\textbf{41.78}}  & \underline{\textcolor{blue}{0.961}}  &\underline{\textcolor{blue}{0.999}} \\ \bottomrule[1.5pt]
\end{tabular}}
\vspace{-5pt}
\caption{\textbf{Ablation on the selection of localized watermark}, where A.T. denotes adaptive watermark transform.}
\vspace{-10pt}
\label{aba1}
\end{table}
\begin{figure}[t!]
    \centering
\includegraphics[width=1\linewidth]{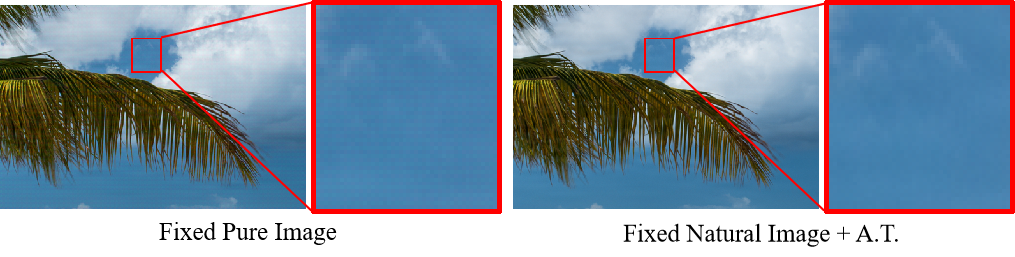}
    \vspace{-20pt}
    \caption{\textbf{Visual comparison of different localized watermarks.} Using a fixed pure localized watermark will cause stripe-like artifacts, but our watermark adaptive transform can alleviate it.}
    \label{fig:aba}
    \vspace{-10pt}
\end{figure}

\subsection{Ablation Study}
\textbf{Exploration on the localized watermark:} Thanks to our hybrid forensic mechanism, we can flexibly choose the localized watermark without compromising localization performance, as reported on Tab.~\ref{aba1}. We find that using a pure color image as a watermark achieves the best localization performance but fails to ensure the watermark's imperceptibility. Specifically, as shown in Fig.~\ref{fig:aba}, hiding a fixed pure image easily produces stripe-like regular artifacts, which greatly compromise watermark security. In contrast, using a fixed natural image significantly alleviates this issue. In case (c), we directly use the original image as the localized watermark, but it does not lead to improvements in either localization performance or fidelity. Based on the natural image, applying our adaptive watermark transformation from Sec.~\ref{ada_watermark} further enhances PSNR by 0.69 dB. More exploration is presented in our \textcolor{blue}{\textbf{supplementary materials}}.

\noindent \textbf{Ablation on the design of deep tamper extractor:} To validate the contribution of our network design, we conducted ablation studies on the degradation query and the tampered image. When the tampered image was removed, leaving only the artifact map as input, the F1-score drops significantly by 0.056. This is because artifacts from degradation can sometimes interfere with those caused by tampering, making the auxiliary input from \( \mathbf{I}_{\text{rec}} \) essential for accurate correction. Furthermore, removing \( \mathbf{Q}_{\text{deg}} \) also reduces the localization precision of the tamper extractor, as it lacked an effective mechanism for fusing \( \mathbf{I}_{\text{rec}} \) and \( \hat{\mathbf{W}}_{\text{loc}} \).

\begin{table}[t!]
\centering
\resizebox{0.95\linewidth}{!}{
\begin{tabular}{lcccc}
\toprule[1.5pt]
Case & Tampered image & Degradation query  & F1 & AUC \\ \hline
(a)  & $\times$  & $\checkmark$ &0.905 &0.962    \\
(b)  & $\checkmark$  & $\times$ &\underline{\textcolor{blue}{0.935}} &\underline{\textcolor{blue}{0.998}} \\ 
(c)  & $\checkmark$ & $\checkmark$  & \textcolor{red}{\textbf{0.961}}  & \textcolor{red}{\textbf{0.999}} \\ \bottomrule[1.5pt]
\end{tabular}}
\vspace{-6pt}
\caption{\textbf{Ablation on two key network designs}, namely the tampered image $\mathbf{I}_{\text{rec}}$ and degradation query $\mathbf{Q}_{\text{deg}}$.}
\vspace{-10pt}
\end{table}

\section{Conclusion}

OmniGuard proposes an innovative hybrid forensic framework that combines proactive watermark embedding with passive extraction. This approach incorporates a deep degradation-aware tamper extractor, adaptive localized watermark transform, and a lightweight AIGC-editing simulator, enabling OmniGuard to achieve superior fidelity, flexibility, and robustness. Experiments show that OmniGuard outperforms state-of-the-art methods by producing higher-fidelity images with fewer artifacts and achieving superior tamper localization and copyright extraction accuracy, even under severe degradations and generative AI edits. As a robust defense against AIGC manipulation, OmniGuard represents a critical advancement for the community, contributing to future developments in copyright protection and digital content authenticity.

{
    \small
    \bibliographystyle{ieeenat_fullname}
    \bibliography{main}
}

\clearpage
\renewcommand{\thesection}{\Alph{section}}
\setcounter{section}{0}
\maketitlesupplementary

\section{Discussions}

\subsection{Limitations and Our Future Works}
The limitations of our method mainly lie in two parts. \textbf{First}, although we have significantly improved OmniGuard's localization accuracy and robustness by introducing a passive detection network, if the degradation is extremely severe and exceeds the robustness threshold of image-in-image steganography, our localization performance will approach that of standard passive detection networks. This issue might be addressed by exploring more advanced steganography frameworks and theories, such as diffusion models~\cite{yu2024cross, xu2024diffusion}, or by establishing relevant evaluation standards to exclude excessively low-quality images from being used for post-event forensics. \textbf{Second}, although the fidelity of our dual watermark exceeds 40 dB, we found that when handling ultra-high-resolution images (\emph{e.g.}, panoramic pictures~\cite{li2024resvr}), the resolution scaling strategy~\cite{bui2023trustmark} will amplify watermark artifacts, which slightly impacts the perceptual quality of our method. However, this is a common issue among all current deep watermarking methods. Therefore, exploring a truly scalable watermarking approach that can handle arbitrary resolutions remains a worthwhile direction for future research.

\subsection{Why Use Joint Training?}
To verify the impact of jointly training localized and copyright watermarks, as opposed to applying two separately trained watermark models to the images, we conducted experiments on both scenarios. We randomly selected 100 images and tampered with them using SD inpainting. For the separate embedding setup, we retrained our original localization watermark embedding and decoding networks and combined them with pre-trained TrustMark~\cite{bui2023trustmark}. The results in Tab.~\ref{joint} clearly demonstrate that without joint training, the watermark fidelity decreases by approximately \textbf{6 dB} PSNR. However, due to the robustness of the passive extraction network, the localization performance remains largely unaffected. This highlights the importance of simultaneously addressing localization and copyright protection in OmniGuard.

\begin{table}[h!]
\centering
\caption{Performance comparison between joint training and using two separate watermarks.}
\vspace{-5pt}
\resizebox{1.0\linewidth}{!}{
\begin{tabular}{lccccc}
\toprule
Method & PSNR (dB) &SSIM & F1 & AUC \\ \midrule
Separate Embedding & 35.46 & 0.966   &0.953  & 0.986             \\ \midrule 
\textbf{Joint Training (Ours)}     & 41.59 & 0.985   &0.975  & 0.999             \\ \bottomrule

\end{tabular}}
\label{joint}
\end{table}

\begin{figure}[t!]
    \centering
    \includegraphics[width=\linewidth]{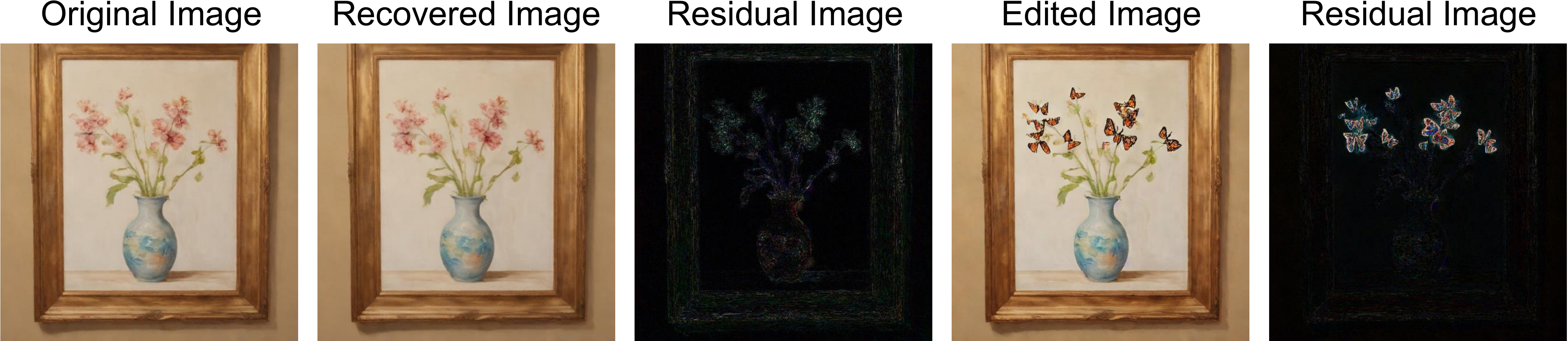}
    \vspace{-15pt}
    \caption{Residual images between the recovered image produced by the VAE and the original image, and between the edited image produced by InstructPix2Pix~\cite{brooks2022instructpix2pix} and the original one.}
    \label{fig:vae}
    \vspace{-10pt}
\end{figure}

\subsection{Why do we choose VAE as surrogate attacks?}
To demonstrate the rationale behind using VAE instead of InstructPix2Pix or other AIGC global editing methods, we visualized two sets of residual maps in Fig.~\ref{fig:vae}. It can be observed that if the image is not processed via a diffusion denoising process and is only encoded and reconstructed using VAE, the artifacts in the residual map are primarily uniformly distributed along the edge information of the original image. When the image is edited using InstructPix2Pix, we find that the error map generated by editing and the error map produced by VAE reconstruction exhibit certain consistency in their distribution. Furthermore, the residual map generated by editing shows smaller differences in areas outside the edited region than the errors observed in the VAE reconstruction. Thus, the distortion caused by VAE on the original image appears to be greater and more global than the distortion introduced by the diffusion process itself. Considering algorithm efficiency and computational resource consumption, we opt to introduce VAE in our training process.

\begin{table*}[h!]
\centering
\caption{Localization performance metrics for EditGuard and OmniGuard under different degradations.}
\vspace{-10pt}
\renewcommand{\arraystretch}{1.2}
\resizebox{0.95\linewidth}{!}{
\begin{tabular}{lccccccccc}
\toprule
Method &Metrics &Clean &JPEG(Q=60) &JPEG(Q=70) &Bri. &Con. &Hue &S.-P. &GS Noise \\ \midrule
\multirow{3}{*}{EditGuard} & F1 & 0.951 &0.515 &0.912 &0.536 &0.876  &0.946  &0.921  &0.821  \\
& AUC  & 0.971  & 0.785  &0.961 &0.817 &0.945 &0.963 &0.968 &0.944    \\
& IoU    &0.935  &0.365    &0.865   &0.410 &0.809 &0.905  & 0.862     &0.709          \\ \midrule
\multirow{3}{*}{\textbf{OmniGuard (Ours)}} & F1 &0.961 &0.810 &0.938 &0.927 &0.926 &0.964 &0.951 &0.958  \\
& AUC  &0.999 &0.982 &0.998 &0.959 &0.960 &0.999 &0.999 &0.999  \\ 
& IoU  &0.928 &0.713 &0.888 &0.999 &0.999 &0.933 &0.911 &0.922  \\  \bottomrule
\end{tabular}}
\label{robust}
\end{table*}

\begin{figure*}[t!]
    \centering
    \includegraphics[width=\linewidth]{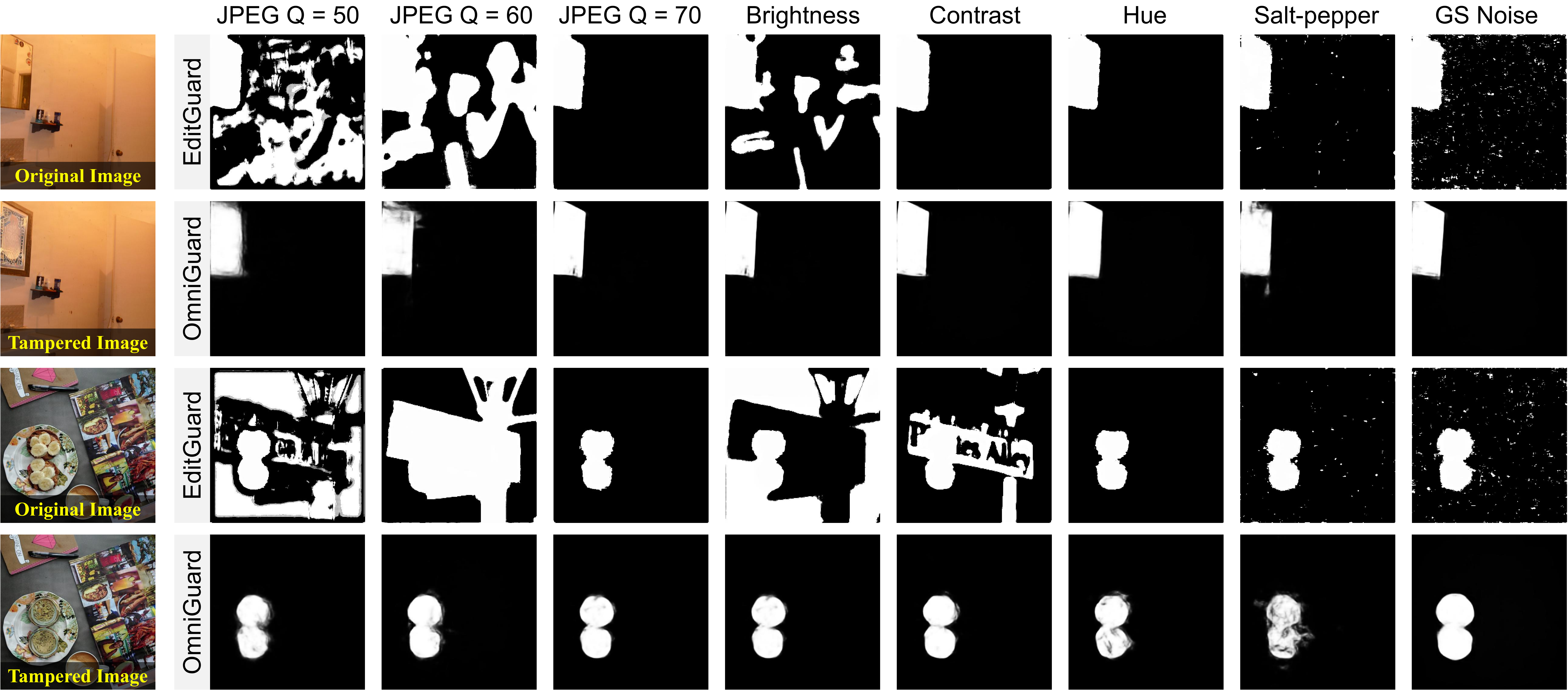}
    \vspace{-15pt}
    \caption{Localization performance of our OmniGuard and EditGuard on several different degradation conditions. Our method can produce clear masks under various noisy conditions, while EditGuard shows confusion and blurriness under certain severe degradations.}
    \label{fig:robustness}
    \vspace{-10pt}
\end{figure*}

\subsection{Exploration on the Localized Watermark}

Considering that the localized watermark has a decisive impact on the final fidelity, we have conducted extensive experiments to identify the optimal localization watermark. Finally, we arrive at the following conclusions:

\begin{itemize}
    \item \textbf{Choice of Color:} Typically, selecting light-colored images allows the steganography network to better hide the localization watermark and achieve higher PSNR. Additionally, using solid-colored images often facilitates subsequent localization and detection.

    \item \textbf{Challenges with Solid Colors:} However, using solid-colored images can result in grid-like or repetitive artifacts on the image, which may be visually unappealing and raise security concerns.

    \item \textbf{Adding Texture:} Adding natural, uncomplicated texture details to solid-colored images significantly improves the fidelity of the watermarked image (such as the light-toned blue sky with clouds used in our paper).

    \item \textbf{Adaptive Watermark Transform:} Coupled with our designed adaptive watermark transform, the fidelity of the hidden localized watermark is further enhanced.
\end{itemize}

These considerations help balance fidelity, detection, and security, making the watermark both effective and visually acceptable.

\section{More Implementation Details}
\textbf{Localized watermark hiding and decoding:} Our localized watermark hiding and decoding network adopts the basic structure of EditGuard~\cite{zhang2024editguard}. It uses a network composed of 16 stacked addition affine transformation layers, where each reversible transformation module employs a DenseBlock. Additionally, we adopt a decoding network constructed with stacked residual blocks to predict the missed high-frequency components \(\hat{\mathbf{z}}\) from \(\mathbf{I}_{\text{rec}}\), enabling accurate inverse decoding of our network.

\textbf{Copyright watermark hiding and decoding:} Following~\cite{bui2023trustmark}, we use a MUNIT-based Unet~\cite{huang2018multimodal} as our watermark embedding network, treating the image watermarking as image translation. The watermark is interpolated to match the original image’s dimension and fused into the input feature via a light network. Our extractor is a standard ResNet50 with the last layer being replaced by a sigmoid-activated FC to predict the watermark. Note that, we use the resolution scaling strategy~\cite{bui2023trustmark} to enable our OmniGuard to support arbitrary resolutions.

\begin{figure*}[t!]
    \centering
    \includegraphics[width=\linewidth]{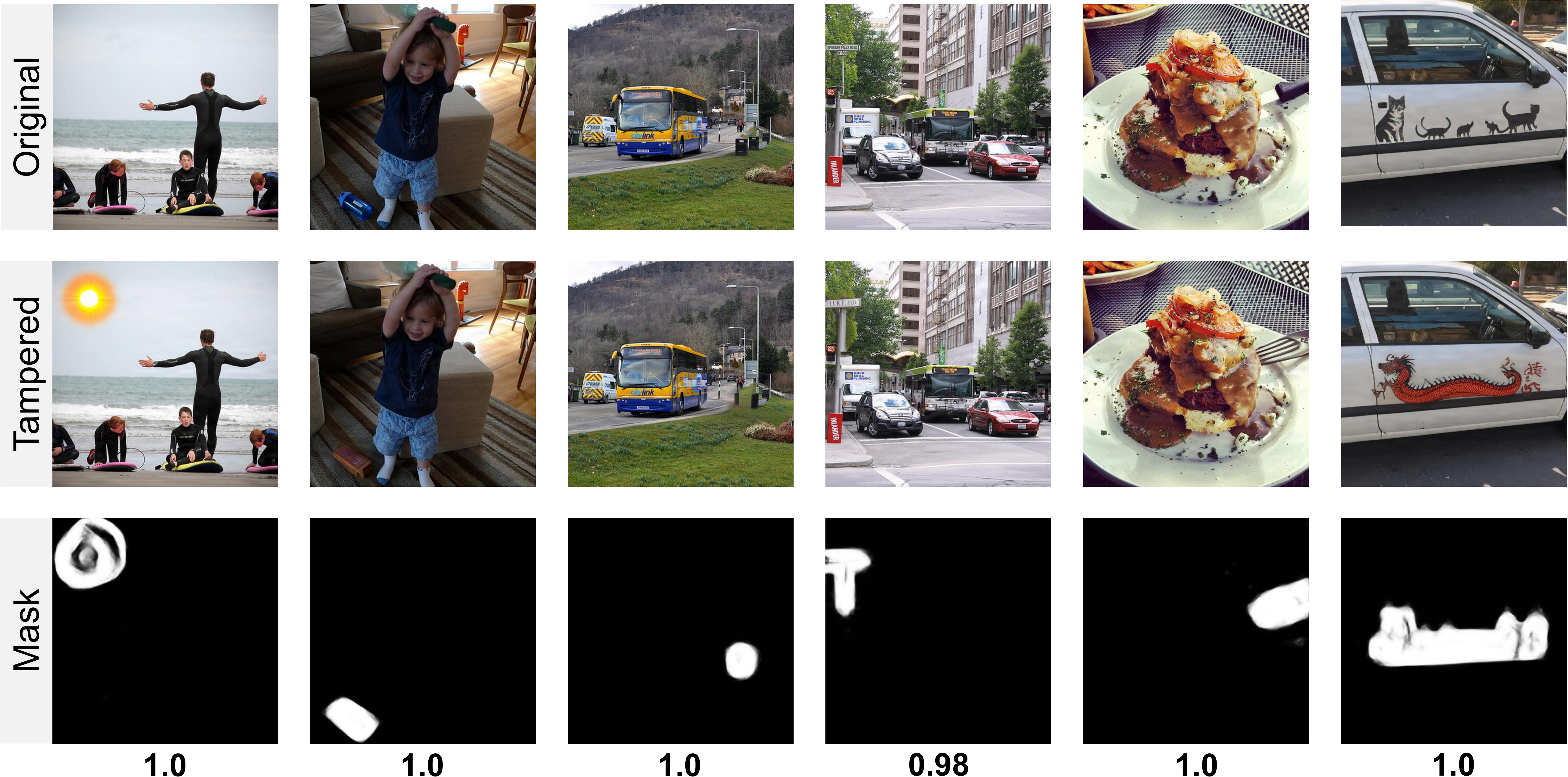}
    \vspace{-5pt}
    \caption{Localization and copyright recovery performance on the most recent AIGC-Editing tool MagicQuill~\cite{liu2024magicquill}. The recovered bit accuracy is shown below. Without any tuning, our method can accurately locate the tampered regions of SOTA editing methods and restore the original copyright.}
    \label{fig:magic}
    \vspace{-10pt}
\end{figure*}

\section{More Experimental Results}

\subsection{Robustness of Our Localization}
To further demonstrate the robustness of our localization performance, we have detailed our localization results under different degradation conditions and compared them with the current state-of-the-art active localization method EditGuard. We selected 1000 images from the COCO dataset and tampered with them using SD Inpaint. We consider various degradation conditions, including JPEG compression (Q=50, 60, 70), Gaussian noise (\(\sigma=15\)), salt-and-pepper noise, and color jitter (adjustments to brightness, contrast, and hue). Tab.~\ref{robust} presents the F1-score, AUC, and IoU of our OmniGuard and EditGuard. We find that under degradation conditions, OmniGuard consistently outperforms EditGuard and remains largely unaffected by different types of degradation. Notably, under severe degradations such as JPEG compression (Q=60) and a 30\% reduction in brightness, OmniGuard shows a significant improvement compared to EditGuard. As shown in Fig.~\ref{fig:robustness}, our method can accurately identify the tampered regions, whereas EditGuard often highlights imprecise and blurry regions under severe degradations.

\begin{figure*}[h!]
\centering
\includegraphics[width=0.9\linewidth]{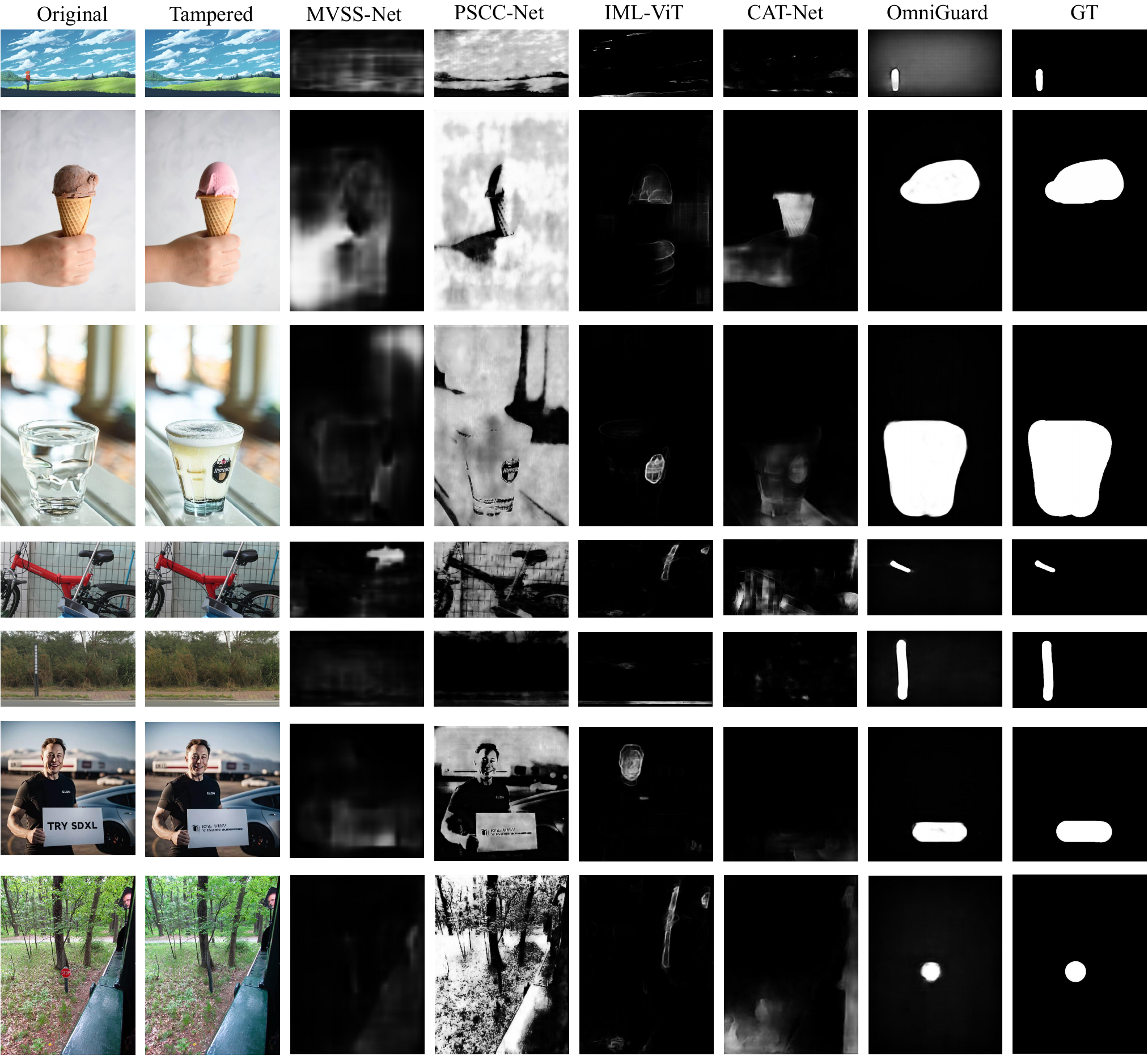}
\vspace{-5pt}
\caption{Localization performance of our OmniGuard and other competitive methods on the SOTA AIGC-Editing Method~\cite{podellsdxl}.}
\label{fig:sdxl}
\label{sdxl}
\end{figure*}

\subsection{Generalization to SOTA AIGC-Edit Methods}
To validate the generalization capability of our method, we tested OmniGuard on two of the latest state-of-the-art AIGC-Edit methods: the recently released and widely discussed MagicQuill~\cite{liu2024magicquill}, and SDXL-inpainting~\cite{podellsdxl}. Fig.~\ref{fig:magic} shows the results of OmniGuard on MagicQuill. It can be observed that our method accurately identifies the tampered regions and correctly extracts the copyright information, even when MagicQuill's edits are highly subtle and difficult to detect with the naked eye. Fig.~\ref{fig:sdxl} further demonstrates that on SDXL, a fine-grained editing method, OmniGuard significantly outperforms passive methods such as PSCC-Net~\cite{liu2022pscc}, MVSS-Net~\cite{dong2022mvss}, and IML-ViT~\cite{ma2023iml} in terms of detection accuracy and generalization. Meanwhile, we can almost completely decode the hidden copyright even under the interference of SDXL inpainting. Notably, when applied to these new AIGC manipulations, OmniGuard requires no fine-tuning or retraining, presenting good generalization ability.

\subsection{Fidelity of Our OmniGuard}
\begin{figure*}[t!]
    \centering
    \includegraphics[width=\linewidth]{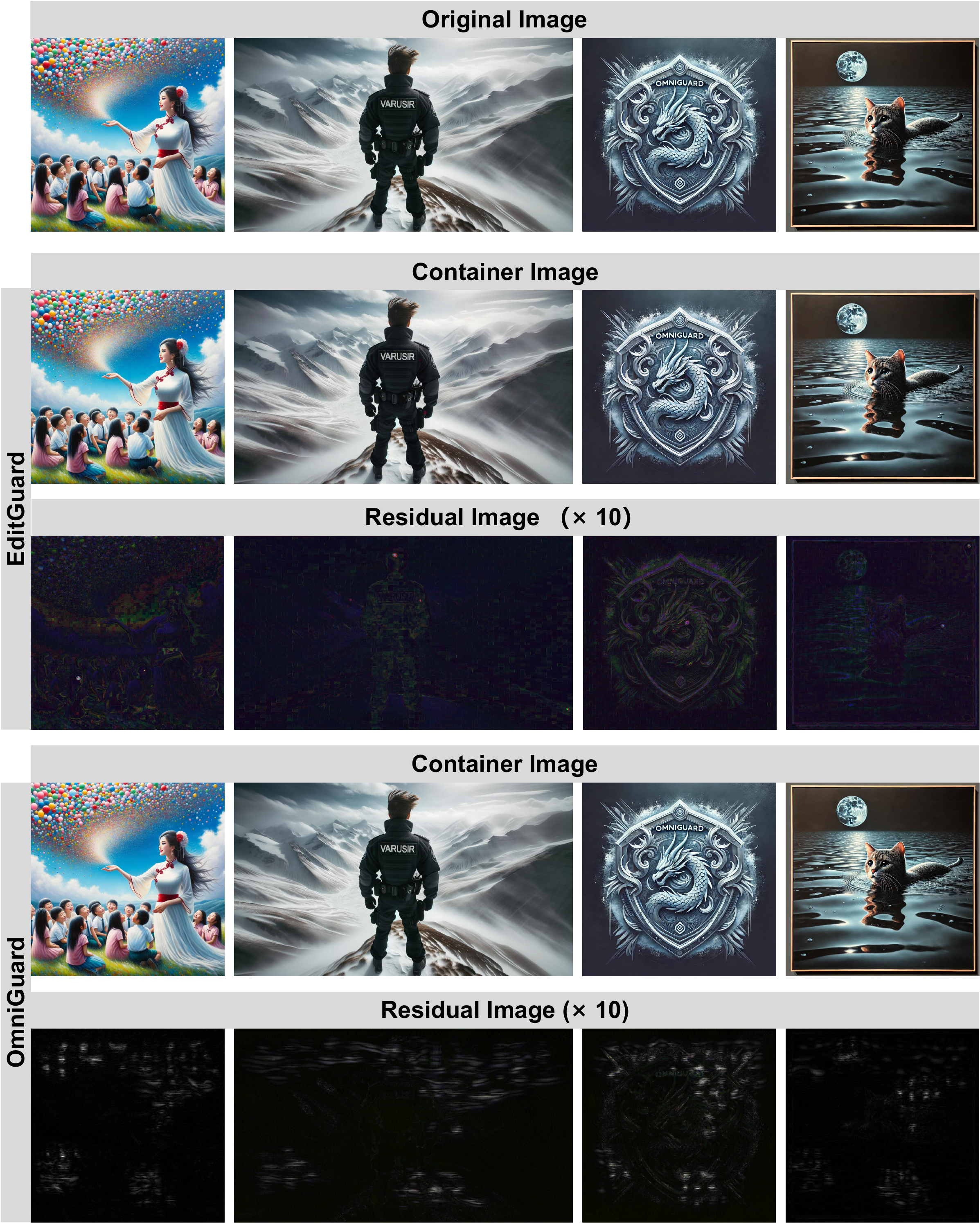}
    \vspace{-15pt}
    \caption{Fidelity comparison between our OmniGuard and EditGuard in some AI-generated high-resolution images. The residual maps, amplified by a factor of 10, are placed below the watermark images. Our OmniGuard shows better fidelity, with watermark artifacts primarily concentrated in background regions that are less perceptible to the human eye.}
    \label{fig:water2}
    \vspace{-10pt}
\end{figure*}

To further validate the fidelity advantages of our method, we test it on high-resolution (1024×1024 and 1792×1024) AIGC-generated images. As shown in Fig.~\ref{fig:water2}, we find that EditGuard, when applied to high-resolution images, tends to produce regular color blocks and artifacts in sparse background areas. In contrast, OmniGuard maintains satisfactory fidelity. We further present residual maps for both methods, scaled by a factor of 10 for better visibility. It can be observed that, overall, OmniGuard’s error map is slighter than EditGuard's while maintaining excellent content adaptiveness. The watermark artifacts are added to areas such as the sky, balloons, clouds, distant mountain peaks, and water ripples in the background. These regions are less perceptible to the human eye compared to the main subjects, such as people and prominent patterns in the image. These results further validate the effectiveness of our method and its general applicability across different data domains.

\end{document}